# Making the case for audience design in conversational AI: Rapport expectations and language ideologies in a task-oriented chatbot

Doris Dippold

*School of Literature and Languages, University of Surrey, Guildford, UK*

d.dippold@surrey.ac.uk

# Making the case for audience design in conversational AI: Rapport expectations and language ideologies in a task-oriented chatbot

**Abstract**

Chatbots are more and more prevalent in commercial and service contexts. They help customers complain about a product or service or support them to find the best travel deals. Other bots provide mental health support or help book medical appointments. This paper argues that insights into users' language ideologies and their rapport expectations can be used to inform the audience design of the bot's language and interaction patterns and ensure equitable access to the services provided by bots.

The argument is underpinned by three kinds of data: simulated user interactions with a chatbot facilitating health appointment bookings, users' introspective comments on their interactions and users' qualitative survey comments post engagement with the booking bot. In closing, I will define audience design for conversational AI and discuss how user-centred analyses of chatbot interactions and sociolinguistically informed theoretical approaches, such as rapport management, can be used to support audience design.

**Introduction**

Chatbots are becoming ever more common in daily life, with an estimated market growth rate of more than 28% annually till 2028 (Grand View Research, 2021). The LITHME (2021) forecast report for example predicts that

> "New generations of chatbots … will not only perform services but also engage in significantly more complex and diverse conversations, including offering advice, thinking through problems, consoling, celebrating, debating and much else. The change here will be in the volume and nature of conversation we hold with technology, and, along with it, our levels of trust, engagement, and even emotional investment." (p. 7)

In this paper, I argue that, in order to ensure user trust of and engagement with chatbots, the bots of the future need to be subjected to 'audience design' to ensure that the bot's language and interaction patterns take into account users' ideologies of language and their rapport expectations. This will ensure that the principles of equality (giving everyone equal opportunities), diversity (valuing difference) and inclusion (ensuring access to services) can be upheld, and AI technologies are accessible to all users.

To give a flavour for my argument regarding the need for audience design, an excursion to existing, non-AI bases studies is warranted. We start with the now famous case study by Gumperz (1982) relating to events at Heathrow airport in the late 1070s. Pakistani and Indians staff there were hired to serve food at a staff canteen but were perceived as rude and uncooperative by supervisors and customers. After recording service interactions in the cafeteria, Gumperz found that the Indian and Pakistani staff tended to use falling information when making service-related requests, e.g., when asking whether the customer wanted gravy, they'd say "Gravy" with falling intonation. This was interpreted as an unnecessary statement and as rude rather than as an offer.

More than 20 years later, Spencer-Oatey & Xing (2003) and Spencer-Oatey (2005) investigated a Chinese-British business meeting. Drawing on transcriptions of video-recordings from the meeting, retrospective comments by participants and researcher observations, the authors showed that interpersonal relationships (rapport) between participants is centred around and can be threatened by individuals' perceptions of different aspects of language and discourse, for example stylistic choices, turn-management, topic choice, sequencing of information etc as well as differences in their assessment of contextual factors.

The third case study now moves us into the world of AI, or conversational AI more specifically. Li and Mao (2015) investigated how the way messages were framed by a virtual health advisory system influenced users' perception of the system and of its value. They were able to show that better alignment of the system's communication style with a user's communication style led to more transparency, enjoyment, informativeness and credibility. Enjoyment of conversation with the humanoid adviser was the best predictor driving intention to ruse the system, followed by perceived transparency and engagement.

These three case studies have one central aspect in common: they all emphasize the power of language and interaction to drive individuals' perception of other social actors, including virtual ones. In this paper, I will take these ideas forward into a proposal for audience design for chatbots.

**Theoretical background**

*Audience design and sociolinguistic theory*

The concept of audience design is derived from Bell's (1984) seminal paper entitled 'Language style as audience design'. Bell rooted his concept explicitly in sociolinguistic theory, defining the factors influencing language variation as both linguistic (e.g., phonology, syntax, etc.) and extralinguistic, including interspeaker (class, age, etc.) and intraspeaker (attention, addressee, topic, etc.) variation. His approach shows reflections of Hymes' (1974) speaking mnemonic which purports that linguistic competence requires an acknowledgement of context in speaking. It also integrates Gumperz (1982)'s notion of contextualisation cue which describes the verbal and nonverbal features which guide speaker interpretation of meaning. Gumperz' example of failed communication at the airport is just one instance in which, through speakers' lack of knowledge about and mastery of key components of 'SPEAKING', namely 'key' as style choices, and 'norms' as the social rules of the event, the interaction had a less than desirable outcome.

Crucially, interactional sociolinguistics research in professional contexts has also revealed that variation in language practices and in perceptions of appropriateness can lead to inequalities, for example in in terms of access of linguistic minorities to the job market (Roberts, 2021). Given the growing importance of chatbots, it is vital that this situation is not mirrored when services are delivered through and accessed through chatbots, and that specific user groups are not excluded from their use. This is particularly important for key public services such as those serving health and welfare purposes: in 2020, research by the King's Fund established that inequalities in access to and experience of health services can arise because "information is not communicated in an easily understandable or culturally sensitive way" (Kings Fund, 2020). Whilst not focusing specifically on digital systems, there is no doubt that these shortcomings can also affect health interventions and services delivered via AI and digital technologies.

This paper will draw on two theoretical concepts which are both framed and broadly informed by sociolinguistic theory, in the sense that they deal with language variation. *Rapport* and *rapport management* on one hand, and *ideologies of language* on the other.

Rapport can be understood as "people's subjective perceptions of (dis)harmony, smoothness-turbulence and warmth-antagonism in interpersonal relations" (Spencer-Oatey & Franklin, 2009), and rapport is managed across three dimensions (Spencer-Oatey, 2008). The first one of these is *face* (the need to have personal qualities and social roles upheld) and *sociality rights and obligations,* consisting of *equity rights* (rights to personal consideration from others) and *association rights* (rights to an association with others that is in keeping with the relationship we have with them). The third dimension is that of *interactional goals*, distinguishing whether speakers have mostly relational or transactional goals. The Chinese-British business meeting discussed by Spencer-Oatey & Xing (2003) and Spencer-Oatey (2005) (see introduction) illustrates what can happen when participants' assessment of these social factors, and consequently their behavioural expectations differ. For example, the participants assessed their own and others' roles and consequently each other's obligations differently.

The second concept, language ideologies, has been defined by Woolard (2020) as both the beliefs about language as well as the language practices indexing these beliefs. In other words, language ideologies are both "mental constructs" and "embodied practices". They are shaped by individuals' social experiences and can be highly normative, as the case studies mentioned in the introduction has shown. However, despite showing that users' behaviours are linked to their beliefs about conversational systems, language ideologies have not yet been an explicit focus of exploration in human-bot interaction, as the following section will show.

*Research insights: Why do we need audience design*

Having defined 'audience design' and rooting it in individuals' expectations of the social world, including their rapport expectations and their beliefs about language, the aim of this section is to provide key research insights into the need for audience design for chatbots and other forms of conversational AI.

The majority of the research on peoples' perceptions of interactions with bots and other forms of conversational AI so far uses experimental methods. Their insights underline not only the importance of building bots or agents which have the ability to build and sustain rapport with their users, but also the importance of aligning the language strategies of the bot with the expectations of their users. For example, Srinivasan (2016) investigated how bots can use different strategies to spur reactions in human users. By systematically manipulating the politeness levels of requests (e.g., positive politeness, negative politeness etc.), they found that the use of positive politeness was more likely to entice people to help the robots. Salem et al. (2014) investigated effects of politeness and culture on robot acceptance and anthropomorphization. They found that the more polite robot was perceived to exhibit greater warmth than the control robot. Moreover, they also found that native speakers of English and of Arabic evaluated the robot differently, suggesting that that differences in language and sociocultural background between users may be guiding users' perceptions. Finally, the importance of rapport building was shown in a study by Bickmore & Pickard (2005) who found that a relational agent supporting users with a new fitness regime attracted more respect, was liked more, and generated more trust than a non-relational agent and made users more likely to want to continue working with the agent.

The importance of alignment of conversational AI systems with user preferences and expectations was also confirmed by Cowan et al. (2016), who operationalised dialectal variation as a factor for users' perceptions of credibility of a navigation system. They found that a navigation system speaking with an Irish voice was considered more credible by its Irish users, no matter how accurate it was. Li & Mao (2015) confirmed the importance of user alignment with their study on a virtual health advisory system. They were able to show that, when the system's communication style aligned with a user's style, it was perceived to be more enjoyable, credible, informative and transparent. Moreover, Følstad et al. (2020), in a questionnaire-based study on user experiences with chatbots, found that user experience, user needs and user expectations varied with age, with younger users expecting more hedonic elements.

Another tranche of studies describes how users interact with conversational AI by drawing on the micro-analytical techniques provided by conversation analysis. For example, Hill et al. (2015) investigated differences between human-human interaction via instant messenger and human-chatbot interaction. They found that users' behaviours were linked to their perceptions of how human-like their conversational partner was. When interacting with chatbots, users used more profanity and less rich vocabulary then when they were interacting with via instant messenger. Braunger et al. (2017)'s study on user interactions with an in-car spoken dialogue system revealed that speakers were influenced by their beliefs about the system, leading them to adapt to the system for example through the use of short concise phrases. Finally, Pelikan (2015) and Pelikan & Broth (2016) investigated how humans adapt to embodied robots, revealing that users adapted their turn design and turn-taking strategies to in line with their expectations of the conversations.

In summary, existing studies suggest that users have more positive perceptions of conversational systems which are perceived to be creating rapport with their users and whose spoken or written output strategies align with user characteristics and expectations. They also show that users' behaviours are linked to their beliefs about these conversational systems, in particular their ideologies of language pertaining to social interaction with and through these systems. In the forthcoming sections, I will explore the role of rapport expectations and language ideologies derived from users' interactions with a task-oriented chatbot. This will later feed into a more extensive discussion of the need for audience design for AI.

**Data**

*Data collection instruments*

The data which support my argument for the need for audience design are drawn from a research project conducted conjointly with start-up company spryt.com. SPRYT have developed an intelligent patient scheduling system which allows patients to schedule medical appointments through Whatsapp via text-based interactions. Patients interact with a digital receptionist – the chatbot – called 'Asa' to schedule appointments, amend or cancel appointments, or respond to a medical screening questionnaire.

This research project represented the first time the scheduling system was tested with users, though this happened in a simulated rather than a real-life clinical setting. Data were gathered through three research instruments:

1. User experience interviews: Ten interviews were planned in collaboration with an experienced UX researcher. Interviews initially centred on users' routines and everyday lives and then their general booking experiences for booking both health-related and other appointments.

After this phase of the interview, the researcher then triggered Asa to make contact with the user via Whatsapp on their phone. Users were given a range of tasks to complete with Asa – e.g., scheduling an appointment, cancelling an appointment, answering the patient invitation questionnaire. With every bot turn, they were asked what they liked about Asa's response, what they disliked, what they thought they should be doing next and what they would actually do next.

2. Questionnaires: After the user experience interviews, spryt.com made some changes to the wording of some turns and to the interaction structure of Asa, based on the researchers' recommendations. Further data were then gathered through a questionnaire which again asked for users' experiences of booking, scheduling, and cancelling appointments and responding to the medical screening questionnaire. Participants completed the questionnaire soon after interacting with Asa. 30 participants were recruited for this stage.

3. User-bot interactions: Across the interview and the questionnaire stage, 36 users made their interactions with the bot available for analysis. In the case of the interactions gathered as part of the interview stage, it is possible to directly cross-match these to participant interviews which allows for insights into specific moments in the interactions. As the questionnaires were anonymous, questionnaires and the interactions of respondents with Asa cannot be cross-matched.

*Participant recruitment and demographics*

Participants were recruited through the researchers' social media channels as well as the university's experimental platform. In the interviews, the majority of participants were UG and PG university students, in addition to two professionals who took part in the research due to professional interest in chatbot development. In the questionnaires, the majority of participants (45%) were between 18 and 24 years old and more than 70% described their ethnicity as 'White'. Therefore, I will use the data only to derive theoretical insights rather than attempting to provide verifiable insights into different user group's perceptions and strategies.

*Data analysis*

The data presented here are drawn from all three data sources. Users embodied language ideologies will be further uncovered through a comparative analysis of user' interaction style in two case studies taken from the interactions gathered in the survey part of the study. Drawing on politeness theory (Brown & Levinson, 1987) and other markers indicating users' orientations towards the bot, such as direct forms of address and discourse markers, this investigation focuses on whether users' orientations are anthropomorphic – i.e., users orient to the agent as humanlike or dehumanized (Waytz et al., 2010). In addition, users' rapport expectations, as expressed in the interviews and questionnaires, will be described with the support of Spencer-Oatey's (2008) rapport management framework. Wherever possible in the context of user statements of rapport gathered through the interviews, these statements will be cross-referenced to users' actual interactions with the bot to derive further evidence on users' ideologies of language.

**Data insights**

*Chatbot interactions: Case Study 1*

Miriam (a pseudonym) was a participant in the second stage of the project. Example (1) is the start of the interaction with Asa, the bot (l. 1-3).

Example (1)

1     *Hi Miriam, I'm Asa your virtual receptionist from University Of Surrey. Here is how I can help. I can*
2     *\*schedule\* appointments, help you \*prepare\* for your appointment or answer a \*question\* you*
3     *may have*
4     Hello, I was wondering if I could book an MRI scan?

Miriam reacts to Asa's introduction by greeting Asa ('hello'), and a request to book an MRI scan. The request is made as a conventionally indirect request, with 'I am wondering' forestalling' the actual request (l. 4).

Example (2)

1    *Sure, what date works for you?*
2    I can do any day next week other than Wednesday
3    *Please pick a time 08:00 09:00 10:00 11:00 12:00 13:00 14:00 15:00 16:00 17:00 18:00 19:00 None*
4    What day?
5    *Did you mean? clinic's \*opening times\* \*hello\* \*prepare\* for appointment \*None\* of these*
6    None of these
7    *Sorry, I didn't quite get that. I'm still learning. Can you please rephrase the date/time you*
8    *provided? Thanks, Asa*
9    Tuesday would be best
10   *Sorry, I didn't understand that. Can you say that in a different way?*
11   Can I have the scan on Tuesday

The second example here is taken from Miriam's booking interaction with Asa. After Asa's prompt to provide a date (l. 1), Miriam indicates her availability, though the only 'named' day within this turn is the day when she is not available (l. 2). This causes Asa to misunderstand Miriam's preferences and prompts her to ask Miriam for her preferred appointment time (l. 3). Given the lack of clarity with regards to the day of the appointment, Miriam reacts by asking a question to confirm the appointment day (l. 4). This request for information is not understood by Asa and prompts a repair request (l. 5), to which Miriam reacts by using one of Asa's prompts from the previous turn (l. 6). Miriam subsequently reacts to Asa's second repair request (l. 7-8) by attempting again to confirm a day of the appointment. Both requests include negative politeness marker, e.g., 'would be' (l. 9) and 'can I have' (l. 11).

Example (3)

1    What do I need to do before the appointment?
2    *You can find us at , Stag Hill, University Campus, Guildford, GU2 7XH, UK*
3    Can I eat before an mri
4    *Are you trying to say? \*general information\* \*schedule\* appointment \*prepare\* for appointment*
5    *\*None\* of these*

| | |
|---|---|
| 6 | Prepare for appointment |
| 7 | *• The scan will be around 30 min • \*Unless we let you know otherwise\*, you can eat and drink as* |
| 8 | *normal and take any prescribed medication • You're welcome to bring a friend or a relative with* |
| 9 | *you • The radiography team with explain the process once you arrive Do you have any other* |
| 10 | *questions or concerns about what to expect on the day?* |
| 11 | Will it hurt |
| 12 | *Sorry, I didn't understand that. Can you say that in a different way?* |
| 13 | Can I cancel the appointment |
| 14 | *Okay,. Please let me know the number of the appointment you would like to cancel: 1. Screening* |
| 15 | *on Tue 22 Mar '22 at 16:00* |

Example (3) represents a point in the interaction at which Miriam is trying to ask questions about the procedure. She starts out by asking a fairly non-specific question about actions to take ahead of the procedure (l. 1). When her initial request is not understood (l. 2), she provides a 'repaired' turn twice (l. 3, 6). Miriam's new information request (l. 11) is not understood ether, prompting Miriam to move out from the information seeking sequence into a cancellation sequence. When making the cancellation request (l. 13), she does so again in an conventionally indirect manner.

*Chatbot interactions: Case Study 2*

John (also a pseudonym) also interacted with Asa during the second stage of the project. Example (4) showcases John's complete booking interaction with Asa.

Example (4)

| | |
|---|---|
| 1 | *Hi John, I'm Asa your virtual receptionist from University Of Surrey. Here is how I can help. I can* |
| 2 | *\*schedule\* appointments, help you \*prepare\* for your appointment or answer a \*question\* you* |
| 3 | *may have* |
| 4 | Schedule |
| 5 | *Sure, what date works for you?* |
| 6 | March 23 |
| 7 | *Okay. you can choose the time of day you want for your appointment: • Morning (8am-12pm) •* |
| 8 | *Afternoon (12pm-5pm) • Evening (5pm-8pm) • Any (8am-8pm)* |
| 9 | Morning |
| 10 | *Please put in your preferred time 08:00 09:00 10:00 11:00 None* |
| 11 | 1000 |
| 12 | *I can offer you these appointment times 08:00 09:00 10:00 11:00 None* |
| 13 | 10:00 |
| 14 | *Hi John. I'm Asa, your virtual receptionist. Happy to confirm your Screening appointment is now* |
| 15 | *booked for Wed 23 Mar '22 at 10:00. Is this ok?* |

After Asa's standard introduction (l. 1-3), John's first turn consist of only one word, indicating his need to schedule an appointment (l. 4). As the interaction progresses, John's subsequent answers are equally short. For example, his date requests include no additional linguistic markers, e.g., politeness markers (e.g., l. 6), and he also uses Asa's own prompts to progress the interaction (l. 9). This is particularly evident when his original date input is not understood by the bot (l. 11) and he subsequently reverts to a time format that is more acceptable to Asa (l. 13). This leads to successful confirmation of the appointment (l. 14-15).

Example (5)

```
1    Prepare
2    • You're welcome to bring a friend or a relative with you • The radiography team with explain the
3    process once you arrive • Please tell us about any special circumstances (for example, if you need
4    wheel-chair access, if you are hard of hearing, if you have poor eyesight or if you are
5    claustrophobic) Do you have any other questions or concerns about what to expect on the day?
6    Directions
7    Here you go…
8    "https://www.google.com/maps/dir/51+Shalford++Shalford,+Guildford+GU48ES+United+
9    Kingdom
10   /+Stag+Hill+University+Campus+Guildford+GU2+7XH+UK"
```

Example (5) is from a sequence in which John asks questions about the preparation for the appointment. He does so without any further politeness markers in a one word turn ('prepare', l. 1). Similarly, he makes his request for directions in only one word, too (l. 6).

Example (6)

```
1    Cancel
2    Okay. Please let me know the number of the appointment you would like to cancel: 1. Screening
3    on Wed 23 Mar '22 at 10:00
4    1
5    Could you tell us why you have cancelled your appointment? Was it related to: 1. Work
6    commitments 2. Transport issues 3. Concerns regarding the appointment 4. Appointment no
7    longer needed 5. Other
8    1
```

When cancelling his appointment, John makes the same stylistic choices as observed previously. He makes a request for cancellation one single word ('cancel', l. 1, and subsequently uses Asa's own prompts to proceed the with the cancellation sequence (l. 4, l. 8).

These examples from John's interaction with Asa show that Miriam and John use rather different approaches to making requests for booking, information and cancellation. Miriam's linguistic behaviour throughout her interaction with Asa, the bot, indicates that she positions Asa as human. Not only does she greet Asa, but she makes the vast majority of her requests with negative politeness markers. These mitigate the force of the request, indicating that the requests might constitute an imposition on the addressee. Hence, Miriam uses similar strategies to those she might use with a human addressee to whom a request might indeed constitute a face-threatening act which limits their freedom of action.

In contrast, John makes each and every request with minimal turns, using just a simple key word. He is also very adapt at using Asa's own prompts when they are provided, a strategy which can be described as 'upwards convergence' or 'accommodation to the bot (Giles & Ogay, 2007). Hence, John uses strategies that are different from what one might expect in face-to-face social interaction, which suggests that he perceives of Asa as dehumanized and thus not worthy of the social conventions that are the norm when speaking to a human addressee.

### *Rapport expectations, rapport threat and beliefs about language*

#### *Face*

'Face' has been described as the positive social value people claim for themselves in social contact (Goffman, 1967). The concept was later extended by Brown & Levinson (1987) in a distinction of positive face (the want for connection) and the want for negative face (the want for distance, lack of imposition). As shown in the example in the introduction, rapport can be threatened when language and interaction patterns do not support these wants.

Example (7)

1   *We want to ask a few simple, medical questions before the scan. Do you have 2-3 min now to*
2   *answer some questions? This is for your safety. Please reply with yes/no*
3   Yes
4   *Thanks. These safety questions take just 3 min, and help us ensure everything is set up correctly*
5   *1.00/0. Do you have a cardiac (heart) pacemaker? Yes/No*
6   No

User comment (1) – relating to example (7):

*I feel this question is pretty like out of the blue is very like cardiac output peacemaker. I think it's quite like it has nothing to do with for why I wanted the MRI. I feel like it's just really random and kind of it doesn't make sense this has nothing to do with your why I contacted them in the 1st place so I don't know why they. It just makes me feel like I don't know why they want me to tell them about this.*

The example here is from the first part of the medical screening questionnaire, with the first question inquiring whether the user has a cardiac pacemaker. The user perceives this as a threat to rapport because of the immediacy of the question and the lack of explanation of its relevance, leading to threat to negative face due to the perceived imposition caused.

Also relating to the medical screening questionnaire was a user comment from the questionnaires:

User comment (2):

*It was easy to do but seemed quite robotic and impersonal. When I answered yes to questions I was expecting Asa to ask me for further information or details but she didn't. It also didn't give me the opportunity to provide further details e.g., what if my answer wasn't just Yes or No. The answers left no space for in between answers/grey areas.*

This user reports perceived threat to rapport on the dimension of positive face because of the bot's perceived 'impersonal' style which don't allow for the want for connection to be satisfied. This is linked specifically to the lack of follow-up questions by Asa on affirmative answers to the medical screening questions, and the fact that the questions and answers did not leave any room for vagueness.

*Sociality rights*

As discussed previously, the *sociality rights and obligations* dimension consist of *equity rights* (rights to personal consideration from others) and *association rights* (rights to an association with others that is in keeping with the relationship we have with them). In addition to rapport threat based on 'face', user comment (2) can also be read as to threaten rapport on accounts of equity rights as the lack of room for elaboration shows a lack of consideration for the users' needs.

Many of the comments in both questionnaires and interviews discuss rapport threat based on perceived equity rights violations, such as when a question is not answered (user comment 3) or the bot displays a lack of understanding (user comment 4):

User comment (3):

*They ignored my question completely and it made me agitated*

User comment (4):

*Too robotic, only understands key words from my phrases, I had to send at least 3 messages for her to understand what I meant.*

Cost-benefit considerations come into play in some user comments:

User comment (5) – relating to example (8):

*For example, I think I don't like the idea for it. I I am just giving a day Tuesday, for example, but I still can't. No exactly which date is available can be the 8th of Tuesday the 8th. It can be 11. Choose day, so, uh, it's a bit overwhelming because I can be available on that day, but I may not be available. May have other responsibilities other days, so I cannot speak exactly at 8. I think I am looking for more freedom to choose the time the date exactly.*

Example (8)

| | |
|---|---|
| 1 | *What date works for you?* |
| 2 | *4/02/2022* |
| 3 | *Our clinic is open Monday through Friday only. Please try specifing a date that does not* |
| 4 | *fall on weekends.* |
| 5 | *Tuesday* |

| 6 | *Please choose a convenient time of day - any - Morning - Afternoon - Evening* |
| 7 | Morning |
| 8 | *Please choose a time slot 09:00 10:00 11:00 None* |
| 9 | 10:00 |

The user here comments on the fact that the booking dialogue does not afford sufficient opportunity for consideration of their needs by the bot. Whilst the bot offers a date, there is no certainty that the user is available at the specific time(s) offered the bot. The user thus felt that their personal freedom to make choices was restricted – an issue also relating to negative face – making the dialogue quite 'costly' in terms of the effort required to achieve a successful booking.

There are also user comments which indicate rapport being threatened based on association rights:

User comment (6):

*Quick and easy. Although I selected option 3 "concerns regarding the appointment" and it didn't prompt me to discuss my concerns or call anyone. That might leave me up in the air as a worried potential patient not sure who to turn to for advice.*

This user descries the bot's lack of response to their indicating that they had concerns about the appointment. These concerns were neither addressed by the bot, nor did a human respond to them in a follow-up call. This goes against this user's perceived rights to an association with a medical service provider who responds to patient concerns.

*Interactional goals*

The rapport dimension of interactional goals is concerned with whether users have a mainly social or a mainly transactional orientation to the bot interactions. In the analysis of interviews 'efficiency' emerged as an important need reported by users, betraying a transactional orientation focused primarily on achieving task goals quickly and efficiently. This is in opposition to another user need described through the theme of 'personalisation'. User comments 7 and 8 and their associated interaction transcripts illustrate these contrasting user needs and goals:

User comment (7) – relating to example (9):

*I think because they just went straight like oh hi and they just meet and they didn't really introduce themselves. I kind of thought like OK this is kind of looked like a robotic responds and also kind of like when they said please choose a convenient time for you. Please choose. It's kind of like. No ones there, it's just the automatic responses. Please choose like I think if it was someone behind the text, I think they'd be more like oh can you or I don't know, it's just something a bit more humanely kind of, not just kind of that direct.*

Example (9):

| | |
|---|---|
| 1 | *Hi Test, I'm Asa from - Asa Clinic, your clinic assistant. Here is how I can help. Just type: \*Schedule* |
| 2 | */ Reschedule / Cancel\* - Book New/Manage existing appointment(s) \*Prep\* - For info on how to* |
| 3 | *prepare for your appointment \*FAQ\* - For our frequently answered questions page \*General\* - For* |
| 4 | *general info (opening times, phone no., directions/maps etc.) \*Admin\* - To speak with someone* |
| 5 | Hello |
| 6 | *Hi Test, Is there anything I can help you with?* |
| 7 | I would like to book an MRI appointment please |
| 8 | *What date works for you?* |
| 9 | Can I book it for the 28th January |

In relation to example (9), the user here discusses their need for a more 'personalised' experience to fulfil core interpersonal / social goals of the encounter, perceiving the dialogue overall as rather 'robotic' and expressing a wish for it to be less 'direct' and involving more direct forms of address, e.g. 'can you'. It is interesting to note that the user here also betrays an overall anthropomorphic orientation to the dialogue, as evidenced by them greeting Asa (l. 5) and their use of politeness strategies in their booking request (l. 7).

In their reflection on the interaction as a whole at the end of the interview, another user displays a much more transactional orientation:

User comment (8):

*In theory this would be great because all I know is that if it was working then I'm not waiting around. I can ask and see what my options are boom and then I have my appointment over and done with without having to wait through a waiting time. And I've also got proof as well over the text message to show that even though they're hopefully will send me a confirmation message. There's also proof via the text that they have texted me saying these are my options and that's the option that I've chosen.*

This user emphasizes primarily the speed and efficiency with which appointments are made by Asa, in particular to booking appointments with the help of a human receptionist. Part of that appreciation for efficiency is also the fact that, given the interaction is recorded in written form, there is a record of the booking.

In addition, the data also revealed instances in which users orient in parallel to social and transactional goals and value both the anthropomorphic and the robotic features of social interaction.

User comment (9):

*Yeah, because I did this kind of like interaction before. I know like the procedure and I know like this is not a real person so the responding speed is it's good. You know they're not like holding back to process the information I type in. So maybe the AI smart. It's smart enough to go through the process. And second, you know the languages for me is. Understandable, it's clear. but however I like, I said before like they don't have like this small talk like they don't have like opening talk that to to make the people who want to book appointment feel welcome, I feel like. but this is it's pretty good. It's pretty good. Yeah, because it's super convenient.*

In comment (9), the user, on one hand, expresses appreciation for the effectiveness and speed by which bookings can be made. On the other hand, they also comment on the fact that, in the earlier stages of the interaction, Asa the bot offers hardly any small talk to make the user feel welcome. This shows that conflicting rapport orientations and language beliefs can co-exist within individuals. These may in turn be reflected within their 'practiced' orientations.

In the next section, I will use the insights gained from the analysis of user interactions with Asa the bot as well as users' reported perceptions in questionnaires and interviews to further strengthen the case for audience design which I started to develop in the introductory parts of this paper. I will also make suggestions how future research can contribute strengthen the agenda for audience design.

**Audience design for chatbots: a way forward**

*Summary*

The data presented above show that, even though the users taking part in this project are from a similar demographic group, their language ideologies – both in form of practiced ideologies and as mental constructs – show considerable variation. They also show that users' practiced ideologies of language are often contrary to what one might expect: users may talk to the bot as if they were a human – using politeness features, referring to the bot by its name – even when the bot does not understand their input and is ineffective in completing the task. Moreover, the data show that, if the bot's language and interaction patterns fail to align with users' rapport expectations, users are likely to have more negative experiences of their interactions. These of course have the potential to disengage users from interacting with bots.

*What is audience design for chatbots and why is it needed?*

Based on these insights, I argue that, in the interest of providing user groups with diverse characteristics with equal access to services which are increasingly being delivered by chatbots, chatbot design needs to take account of these expectations through audience design. I define audience design as follows:

> *Audience design for chatbots is the adaptive design of language and interaction patterns in conversational agent output to accommodate to different user groups' rapport needs and expectations about language.*

Engaging in audience-design would go some way towards meeting Microsoft's (2019) guidelines for human-AI interaction design, which include "Match relevant social norms: Ensure that experience is delivered in a way that users would expect, given their social and cultural context" and "Mitigate social biases: Ensure the AI system's language and behaviours do not reinforce undesirable and unfair stereotypes and biases". It will also enhance the user-centredness of chatbots, which has been a frequently bemoaned issue in relation to conversational AI. For example, Kopp & Krämer (2021) purport that chatbots are currently designed in a technology- rather than user-centred manner, "based on the assumption that all necessary reasoning for a suitable, self-contained communicative response can be done on-the-spot and based on a sufficiently large set of training data" (Kopp & Krämer, 2011, p 2).

Conversely, in their report on 'Re-humanizing artificial intelligence, Hantula et al. (2021) ask that AI development take into account human needs and desires:

> *"People are sensual flickering creatures with changing needs and aspirations. Their situations and circumstances vary. They have feelings. Sometimes they act rationally but often not. [...] AI-powered systems for everyday use [...] should accept that the same principles which brilliantly work when building algorithmic systems for diagnosing rare diseases or modelling the changes in the global climate patterns, for example, cannot be applied here. [....] Developers of algorithmic systems [...] ought to start experimenting with more multi-layered algorithmic systems what will be better at adjusting themselves in people's daily alteration" (p. 43).*

The view held here of a multi-layered approach to AI design is supported by other studies. For example, Sutton et al.'s (2019) study on sociophonetic design strategies inhuman/computer interaction. Focusing on voice-user interfaces such as Siri and Amazon Alexa, they argue that the design of VUI voices needs to take into account peoples' experiences and the speech ideologies they hold, and that the fields of human-computer interaction and sociophonetics should thus be further integrated through an approach that "aims to address people's desires, values and feelings through enriching interacting with technology by considering the personal narratives that they bring to an interaction." (p. 11). Van der Goot (2021) describe customers' 'communication journeys' with chatbots through the help of interviews with respondents of different genders, ages and educational levels. They found that customer journeys were strongly influenced by customer' prior expectations of the interaction, their experiences during the conversations, the effort required and the extent to it which the bot appeared to be human-like. Finally, the study also revealed that there were differences in users' expectations of language style, e.g. of the use of formal vs. informal style. This suggesting that these expectations and experiences need to be taken into account by audience design.

*A research agenda for audience design*

When the conversational pathways of chatbots are scripted and defined, this is usually done with the support of conversation designers. The Conversation Design Institute, which offers courses and certifications in Conversation design, defines the discipline as "the practice of making AI assistants more helpful and natural when they talk to humans. It combines an understanding of technology, psychology, and language to create human-centric experiences for chatbots and voice assistants" (Van Dam, 2020).

Based on this definition as well as the theoretical review and the data presented earlier, I purport that audience design should be seen as an aim as well as an outcome of conversation design. As an aim, audience design is able to support principles of inclusive design, defined in a recent monograph on conversation design (Deibel & Evanhoe, 2021) as "design that considers the full range of human diversity with respect to ability, language, culture, gender, age and other forms of human difference" (p. 270).

As an outcome, effective audience design needs to be underpinned by relevant empirical methods and theoretical concepts. As the literature review above has shown, the methods in research on conversational AI are still dominated by experimental approaches whose results are evaluated by statistical / quantitative means. Ethnographic approaches which investigate user interactions with AI, and investigations which focus on user perceptions do exist, but they also need to be rolled out more systematically to different bot interaction contexts (text-based vs. spoken voice bots, transactional vs. social bots) and, crucially, move out of simulated environments into 'the wild', i.e. life contexts.

The same overall picture also applies to UX (user experience) research which conversation design is informed by and part of. UX research can be conducted from a practice-led perspective with the aim of deriving specific recommendations for product design, as well as academic perspectives aiming to reveal "how different users interact with technologies (and each other, and how cultural and societal factors come into play in these interactions" (Girucci & Pruchiniewska, 2022, p. 28). However, common methods in UX research are not well suited to exploring these societal factors. Robinson et al.'s (2017) review of 400 UX research studies published between 2000-2016 found that, whilst UX methods were diverse, the field was dominated by as questionnaires and interviews. In contrast, studies focusing on the wider user context, e.g., through ethnography, were rare.

As our data have shown, any research which has the purpose of underpinning audience design for chatbots will benefit strongly from making the conversations themselves centre stage as their analysis can provide insights into users' practiced ideologies of language, and potentially differences between different user groups' ideologies. A study by Kvale et al. (2020) for example shows that "successful chatbot conversations depend on the user, the chatbot, and the service provider" (p. 198). In particular, they found consumers' linguistic behaviours – for example the way they expressed their needs – to have an influence on the success of chatbot dialogues. Focusing on user ideologies of language as expressed through their stylistic choices will also ensure that research perspectives in human-bot interaction is not skewed towards the bot but widen the analytical gaze towards users and the contexts they 'bring along' to an interaction.

For the same reason, research will also benefit from a strong theoretical underpinning by theories of language and language variation to add weight to existing frameworks explaining used to explain trust in machines, such as for example the ABI framework (Ability, Benevolence, Integrity) (see Toreini et al., 2019). Whilst *ability*, in this framework, would be primarily related to describing users' abilities to complete a task successfully with the help of the conversational system, *benevolence* has strong resonance in the *face* and in the *sociality rights* component of rapport management as perceived violations of rapport have the potential to make the user doubt the benevolence of the system. Finally, *integrity* shows links to users' ideologies of language, indicating whether they user finds the bot's language and interaction patterns acceptable.

Audience design can of course also benefit from the experimental and quantitative approaches that have so long dominated research on chatbots and conversational AI. These approaches can combine with the qualitative frameworks of analysis mentioned above to investigate differences across user groups and influence audience design and perhaps inform the co-design of conversational AI (Steen, 2013) in this way.

**Conclusion**

This paper has primarily been a theoretical one: it has developed the concept of 'audience design' for chatbots and other forms of conversational AI, arguing that careful audience design allows AI-based social interaction to take into account the needs of specific user groups, moving away from a "one size fits all" approach. Through excerpts from user interactions with AI, user interview and surveys I have also shown that the specific theoretical lenses of ideologies of language – in particular users' practiced language ideologies – and rapport management can provide an ideal underpinning for empirical studies supporting audience design. With commitment to audience design, conversational AI can be developed in a people-centric way, increasing user engagement with and enjoyment of bots and related technologies and provide equitable access to services delivered through these technologies.

**References**

Bell, A. (1984). Language style as audience design. Language in Society 13 (2), 145-204. https://doi.org/10.1017/S004740450001037X.

Bickmore, T., Picard, R. (2005). Establishing and maintaining long-term human-computer relationships. *ACM Transactions on Computer-Human Interaction* 12(2), 293-327. https://doi.org/10.1145/1067860.1067867.

Braunger, P., Maier, W., Wessling, J. & Werner, S. (2017). Natural language input for in-car spoken dialog systems: how natural is natural? *Proceedings of the SIGDIAL 2017 Conference,* 137-146. 10.18653/v1/W17-5517.

Brown, S. & Levinson, P. (1987). *Politeness. Some Universals in Language Usage.* Cambridge: Cambridge University Press.

Cowan, B.R., Gannon, D., Walsh, J., Kinneen, J., O'Keefe, E., & L. Xie (2016). Towards Understanding how Speech Output Affects Navigation System Credibility. *Proceedings of CHI 2016,* 2805–2812. https://doi.org/10.1145/2851581.2892469

Deibel, D. & Evenhoe, R. (2021). *Conversations with things: UX Design for Chat and Voice.* Rosenfeld.

Følstad, Asbjørn & Brandtzæg, Petter (2020). Users' Experiences with Chatbots: Findings from a Questionnaire Study. Quality and User Experience 5(3), 10.1007/s41233-020-00033-2.

Kopp, S. & Krämer, N. (2021). Revisiting human-agent communication: the importance of joint co-co-construction and understanding mental states. *Frontiers in Psychology* 12. https://doi.org/10.3389/fpsyg.2021.580955.

Kvale, K., Sell, O., Hodnebrog, S. & Følstad, A. (2020). Improving conversations: Lessons learnt from manual analysis of chatbot dialogues. In. Følstad, A. et al. (eds.), *Conversations 2020* (pp. 187-200). Springer. https://doi.org/10.1007/978-3-030-39540-7_13.

Li, M. & Mao, J., (2015). Hedonic or utilitarian? Exploring the impact of communication style alignment on user's perception of virtual health advisory services. *International Journal of Information Management* 35 (2), 229–243. https://doi.org/10.1016/j.ijinfomgt.2014.12.004.

LITHME (2021). The Dawn of the Human-Machine Era. A Forecast of new and Merging Language Technologies. https://doi.org/10.17011/jyx/reports/20210518/1.

Microsoft (2019). Guidelines for human-AI interaction design. https://www.microsoft.com/en-us/research/blog/guidelines-for-human-ai-interaction-design/.

Pelikan, H. (2015). How humans adapt to a robot recipient: an interaction analysis perspective on human-robot interaction. Bachelor Thesis: University of Osnabrück. https://www.diva-portal.org/smash/get/diva2:823843/FULLTEXT01.pdf.

Pelikan, H. & Broth, M. (2016). Why that Nao? How humans adapt to a conventional humanoid robot in taking turns-at-talk. *CHI '16: Proceedings of the 2016 CHI Conference on Human Factors in Computing Systems,* 4921-4932. https://doi.org/10.1145/2858036.2858478.
27